\documentclass[conference]{IEEEtran}
\usepackage{cite}
\usepackage{amsmath,amssymb,amsfonts}
\usepackage{algorithmic}
\usepackage{graphicx}
\usepackage{textcomp}
\usepackage{xcolor}
\def\BibTeX{{\rm B\kern-.05em{\sc i\kern-.025em b}\kern-.08em
    T\kern-.1667em\lower.7ex\hbox{E}\kern-.125emX}}
    
\usepackage[breaklinks=true,colorlinks,bookmarks=false]{hyperref}
\usepackage[nameinlink]{cleveref}
\usepackage[ruled,vlined]{algorithm2e} 
\usepackage{subfigure}

\usepackage{comment}
\usepackage{todonotes}

\crefname{algocf}{algo.}{algorithms}
\Crefname{algocf}{Algorithm}{Algorithms}
\Crefname{equation}{}{Eqs.}
\Crefname{figure}{Fig.}{figure}
\Crefname{section}{Sec.}{section}

\newcommand{\auroc}{\mathit{AUROC}}
\newcommand{\auc}{\mathit{AUC}}
\newcommand{\acc}{\mathit{ACC}}

\begin{document}

\title{False Negative Reduction in Video Instance Segmentation using Uncertainty Estimates\\
}

\makeatletter
\newcommand{\linebreakand}{%
  \end{@IEEEauthorhalign}
  \hfill\mbox{}\par
  \mbox{}\hfill\begin{@IEEEauthorhalign}
}
\makeatother

\author{\IEEEauthorblockN{Kira Maag}
\IEEEauthorblockA{\textit{University of Wuppertal, Germany} \\
kmaag@uni-wuppertal.de}
}

\maketitle

\begin{abstract}
Instance segmentation of images is an important tool for automated scene understanding. Neural networks are usually trained to optimize their overall performance in terms of accuracy. Meanwhile, in applications such as automated driving, an overlooked pedestrian seems more harmful than a falsely detected one. In this work, we present a false negative detection method for image sequences based on inconsistencies in time series of tracked instances given the availability of image sequences in online applications. As the number of instances can be greatly increased by this algorithm, we apply a false positive pruning using uncertainty estimates aggregated over instances. To this end, instance-wise metrics are constructed which characterize uncertainty and geometry of a given instance or are predicated on depth estimation. The proposed method serves as a post-processing step applicable to any neural network that can also be trained on single frames only. In our tests, we obtain an improved trade-off between false negative and false positive instances by our fused detection approach in comparison to the use of an ordinary score value provided by the instance segmentation network during inference.
\end{abstract}

%
%
%
\section{Introduction}
Instance segmentation combines object detection which means the task of categorizing as well as localizing objects using bounding boxes, and semantic segmentation, i.e., the pixel-wise classification of image content. In instance segmentation, the localization of objects is performed by labeling each pixel that corresponds to a given instance, see \Cref{fig:gt_pred}. Thereby, instance segmentation provides precise information about the most important classes of instances. State-of-the-art approaches are mostly based on convolutional neural networks. Neural networks as statistical models produce probabilistic predictions prone to error, for this reason, it is necessary to understand and minimize these errors. In safety critical applications like automated driving \cite{Le2018} and medical diagnosis \cite{Ozdemir2017}, the reliability of neural networks in terms of uncertainty quantification \cite{Gal2016} and prediction quality estimation \cite{DeVries2018,Maag2019,Maag2020} is of highest interest.
\begin{figure}[t]
\center
    \includegraphics[width=0.485\textwidth]{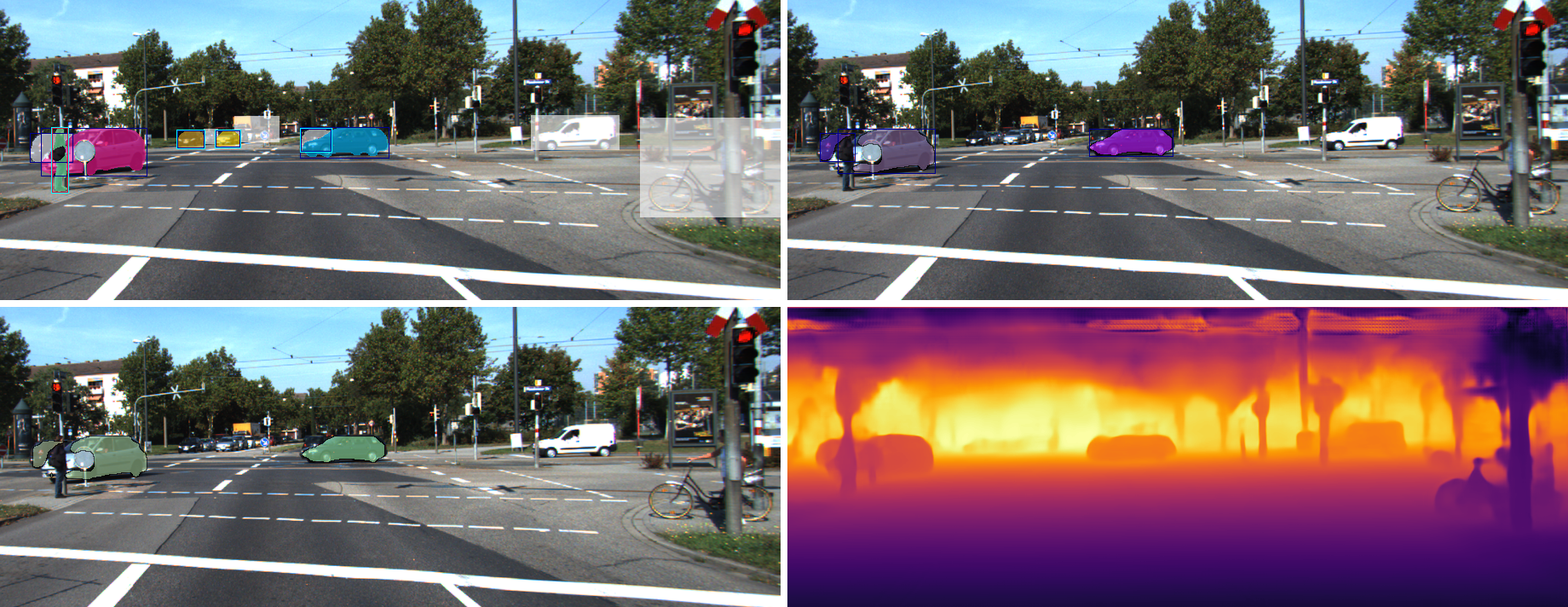}
    \caption{\emph{Top left}: Ground truth image with ignored regions (white). The bounding boxes drawn around the instances represent the class, red denotes pedestrians and blue cars. The cyan colored bounding boxes highlight non-detected instances. \emph{Top right}: Instance segmentation. \emph{Bottom left}: A visualization of the calculated instance-wise IoU of prediction and ground truth. Green color corresponds to high IoU values. \emph{Bottom right}: Depth estimation map.}
    \label{fig:gt_pred}
\end{figure}
Instance segmentation networks (for example Mask R-CNN \cite{He2017} and YOLACT \cite{Bolya2019}) provide for each object a confidence value, also called score. However, these scores do not correspond to a well-adjusted uncertainty estimation \cite{Guo2017} as they can have low values for correctly predicted instances and high values for false predictions. This problem is addressed by confidence calibration \cite{Kueppers2020} where the confidence values are adjusted to improve the prediction reliability. During inference of an instance segmentation network, a score threshold is applied to remove all instances with low confidences. This is done to balance the number of false positive and false negative instances. Nevertheless, this can result in correctly predicted instances vanishing as well as many false positives remaining. These errors shall be reduced to improve network performance in terms of accuracy. In applications such as automated driving, the detection of road users, i.e., the reduction of false negative instances, is particularly important. In other words, it is preferable to predict road users incorrectly to missing them. For this reason, we use a relatively small score threshold value during inference and apply a light-weight false negative detection algorithm on these remaining instances to find possible overlooked ones. As the number of instances can be greatly increased by our algorithm to reduce false negatives, we use a false positive pruning based on uncertainty estimates. We utilize this fused approach to improve the networks' performance and to reduce especially false negatives, i.e., attain a high recall rate, compared to using ordinary score thresholds.

In this work, we introduce a false negative reduction method based on uncertainty estimates for instance segmentation networks. Our approach serves as a post-processing step applicable to any network. First, the predicted instances obtained by a neural network are tracked in consecutive frames. Next, our light-weight detection algorithm is applied which is based on inconsistencies in the time series of the tracked instances such as a gap in the time series or a sudden end. We detect these cases and construct new instances that the neural network may have missed using the information of previous frames. To this end, we create time series of the instances and shift the pixel-wise mask of a previous frame to the predicted current position of the new instance via linear regression. By this detection method, the number of instances can be greatly increased and thus, we deploy meta classification to improve also the precision rate. Meta classification for instance segmentation is presented in \cite{Maag2020} and was previously introduced for semantic segmentation in \cite{Rottmann2018}. Meta classification addresses the task of predicting if a predicted instance intersects with ground truth or not. To quantify the degree of overlap between prediction and ground truth, we consider a commonly used performance measure, the intersection over union (IoU) \cite{Jaccard1912}. In object detection, meta classification refers to the task of classifying between IoU $< 0.5$ and IoU $\geq 0.5$ and in semantic segmentation between IoU $=0$ and IoU $>0$. Since instance segmentation is a combination of both, we classify between IoU $< h$ (false positive) and IoU $\geq h$ (true positive) for all predicted instances using different thresholds $h$ between $0$ and $0.5$. We use meta classification as false positive pruning after the application of our detection algorithm to improve the overall network performance in comparison to score thresholding during inference. As input for the meta classification model, instance-wise metrics are defined. These metrics characterize uncertainty and geometry of a given instance like instance size, geometric center and occlusion level. In addition, we apply a depth estimation network which can infer in parallel to the instance segmentation network. Based on the resulting heatmap, we construct further metrics aggregated per instance. We complete our set of metrics with a few measures presented in \cite{Maag2020} that are based on expected position, changes in the shape and survival time analysis of instances in an image sequence. 

In this work, we present a post-processing method for performance improvement and in particular, for false negative reduction based on uncertainty estimates. We only assume that image sequences of input data and a trained instance segmentation network are available. In our tests, we employ two instance segmentation networks, YOLACT and Mask R-CNN, and deploy these networks to the KITTI \cite{Geiger2012} and the MOT \cite{Milan2016} dataset for multi-object tracking and instance segmentation. The source code of our method is publicly available at \url{http://github.com/kmaag/Temporal-False-Negative-Reduction}. Our contributions are summarized as follows:
\begin{itemize}
    \item We introduce a light-weight detection algorithm for predicted instances obtained by a neural network and tracked by the algorithm of \cite{Maag2020}. Furthermore, the time-dynamic metrics which serve as input for the meta classification model are defined.
    \item We study the properties of the metrics and the influence of different lengths of time series which are used as input for the meta classification model. We perform meta classification to detect false positive instance achieving $\auroc$ values of up to $99.30\%$.
    \item For the first time, we demonstrate successfully that a post-processing false negative detection method can be traded for instance segmentation performance. We compare our approach with the application of a score threshold during inference. Our detection algorithm fused with uncertainty based meta classification achieves area under precision-recall curve values of up to $95.39\%$.
\end{itemize}
The paper is structured as follows. The related work on false negative reduction methods is discussed in \Cref{sec:rel_work}. In \Cref{sec:method}, we introduce our method including the detection algorithm, instance-wise metrics and meta classification. The numerical results are presented in \Cref{sec:result}. We conclude in \Cref{sec:concl}.
%
%
%
\section{Related Work}\label{sec:rel_work}
In this section, we present the related work on methods for false negative reduction, i.e., recall rate increase. 
For semantic segmentation, a method to achieve a higher recall rate is proposed in \cite{Xiang20190} based on the loss function, classifier and decision rule for a real-time neural network. The similar approach in \cite{Xiang20191} uses an importance-aware loss function to improve the networks' reliability. In \cite{Chan2020}, the differences between the Maximum Likelihood and the Bayes decision rule are considered to reduce false negatives of minority classes by introducing class priors which assign larger weight to underrepresented classes.
The following methods address false negative reduction for the object detection task. In \cite{Xiao2003}, a boosting chain for learning successive boosting is presented. Using previous information during cascade training, the model is adjusted to a very high recall rate in each layer of the boosting cascade. An ensemble based method is proposed in \cite{Casado2020} where the number of false negatives is reduced by the different predictions of the networks, i.e., some objects that are not detected by one network could be detected by another one. In addition to false negative reduction, the number of false positives is also decreased in \cite{Goswami2020} by training a neural network with differently labeled images composed of correct and incompletely labeled images. In \cite{Wang2007}, a set of hypotheses of object locations and figure-ground masks are generated to achieve a high recall rate and thereafter, false positive pruning is applied to obtain a high precision rate as well. 
For $3$D object detection, a single-stage fully-convolutional neural network is introduced in \cite{Yang2018}. Instead of using a proposal generation step, the model outputs pixel-wise predictions where each prediction corresponds to a $3$D object estimate resulting in a recall rate of $100\%$.
In one work \cite{Li2017} applied to instance segmentation, a high recall rate is ensured by generating $800$ object proposals for any given image, followed by the application of the maximum a-posteriori probability principle to produce the final set of instances. In \cite{Zhou2020}, a recurrent deep network using pose estimation is considered to improve instance segmentation in an iterative manner and to obtain a high recall rate. Another approach \cite{Lin2020} constructs a variational autoencoder on top of the Mask R-CNN to generate high-quality instance masks and track multiple instances in image sequences. To reduce false negative instance predictions, spatial and motion information shared by all instances is captured. The method presented in \cite{Maag2020} compares the ordinary score thresholding during inference of an instance segmentation network with meta classification as a post-processing step. Instead of using a score threshold, meta classification, i.e., false positive detection, is used to improve the trade-off between the number of false positive and false negative instances.

In comparison to most of the described false negative detection approaches, our method does not modify the training process or the network architecture, instead our detection algorithm serves as post-processing step applicable to any instance segmentation network. The work closest to ours, \cite{Maag2020}, compares meta classification applied to all predicted instances obtained by a neural network with the use of a score threshold during inference. However, the number of false negative instances detected are limited to those predicted by the network without using a score threshold, since no other instances are generated. We go beyond this method and present a detection algorithm that generates additional instances to the predicted ones and thus, instances missed by the network can be detected.
%
%
%
\section{Method}\label{sec:method}
In instance segmentation as an extension of object detection, instances represented as pixel-wise masks are predicted with corresponding class affiliations. During inference, a score threshold is used to remove instances with low scores followed by a non-maximum suppression to avoid multiple predictions for the same instance. On these remaining instances, we apply the tracking algorithm for instances introduced in \cite{Maag2020} to obtain time series.
Our false negative detection method is based on temporal information of tracked instances. We detect inconsistencies in the time series and construct new instances in these cases. As a result, the number of instances can be greatly increased and thus, we deploy meta classification (false positive detection) to filter them out. To this end, we present meta classification which uses time-dynamic metrics aggregated on instances as input. Our method is intended to improve the networks' performance in terms of accuracy. An overview of the method is shown in \Cref{fig:method}.
\begin{figure}[t]
\center
    \includegraphics[width=0.485\textwidth]{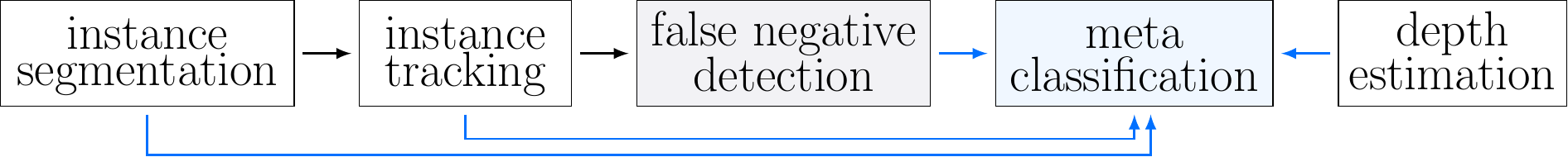}
    \caption{Overview of our approach which consists of false negative detection and false positive pruning applied to tracked instances. For metrics construction, information is extracted from instance geometry, instance tracking and depth estimation (\emph{blue arrows}). The resulting metrics serve as input for the meta classification.}
    \label{fig:method}
\end{figure}
%
%
\subsection{Detection Algorithm}\label{sec:detection}
In this section, we introduce our detection method to identify possible non-detected instances in image sequences. We assume that an instance segmentation is available for each frame, as our method serves as a post-processing step and is independent of the choice of instance segmentation network. An example of an instance segmentation with corresponding ground truth image is shown in \Cref{fig:gt_pred}. The white areas within the ground truth image are ignored regions $\mathcal R$ with unlabeled instances, here cars and pedestrians. Besides, each instance $i$ of an image $x$ has a label $y$ from a prescribed label space $\mathcal{C}$. All predicted instances where $80\%$ of their number of pixels are inside an ignored region are not considered for detection and further experiments as an evaluation for these instances is not feasible. This calculated ratio, i.e., the overlap of an instance $i$ with an instance or region $j$, is formulated through
\begin{equation}
    O_{i,j} = \frac{| i \cap j |}{| i |} \, .
\end{equation}
Given an image $x$, the set of predicted instances not covered by ignored regions is denoted by $\hat{\mathcal I}_x$. 
In addition to instance segmentation, we also assume a tracking of these instances in the image sequences. To this end, we use the tracking algorithm for instances presented in \cite{Maag2020}. Instances are matched according to their overlap in consecutive frames by shifting instances based on their expected location in the subsequent frame. The overlap of two instances $i$ and $j$ is given by
\begin{equation}\label{eq:overlap}
    \tilde{O}_{i,j} = \frac{| i \cap j |}{| i \cup j |} \, .
\end{equation}
The tracking algorithm is applied sequentially to each frame $\{ x_{t} : t=1,\ldots,T \}$ of an image sequence of length $T$. For a description of how an instance $i \in \hat{\mathcal I}_{x_{t-1}}$ in frame $t-1$ is matched with an instance $j \in \hat{\mathcal I}_{x_{t}}$ in frame $t$, we refer to \cite{Maag2020}.

Our detection method is based on inconsistencies in the time series of the tracked instances such as a gap in the time series or a sudden end. We detect these cases and construct new instances that may have been overlooked by the neural network using the information of previous frames. To this end, time series of the geometric centers of the instances are created and the pixel-wise mask of an instance of a previous frame is shifted to the predicted position of the new instance using a linear regression. The geometric center of instance $i$ in frame $t$ represented as pixel-wise mask is defined by 
\begin{equation}\label{eq:geom_center}
    \bar{i}_{t} = \frac{1}{|i|} \sum_{(z_{v}, z_{h}) \in i} (z_{v}, z_{h})
\end{equation}
where $(z_{v}, z_{h})$ denotes the vertical and horizontal coordinate of pixel $z$. If $t^{\text{last}} < t$ is the last frame in which instance $i$ occurs, then we adopt this pixel-wise mask as the representation for the new instance in frame $t$. To avoid false positives, we track a lost instance for at most one second ($10$ frames) and check if the instance is mostly covered by another one or an ignored region. A detailed description of our detection method is depicted in \Cref{alg:detect}.
\begin{algorithm}[!th]
  \caption{False negative detection algorithm}
  \SetAlgoLined
  \DontPrintSemicolon
  \SetAlgoNoEnd
  $ \hat{\mathcal I}^{\text{detect}}_{x_{t}} := \{ \} \ \forall \ t=1,\ldots,T $ \;
  /* detect instances  */ \;
  \For{$i \in \bigcup_{t=1}^{T} \hat{\mathcal I}_{x_{t}}$}{
    $G := \{ \} $ /* time series of geometric centers */ \;
    $t^{\text{last}} := 0 $ /* last previous frame in which instance $i$ occurs  */  \;
    \For{$t = 1,\ldots, T$}{
        \If{$i$ exists in frame $t$}{
            $G := G \cup \{ \bar{i}_{t} \} $ \;
            $t^{\text{last}} := t $ 
        }
        \If{$i$ does not exist in frame $t$ and $t-t^{\text{last}} \leq 10$ and $|G| \geq 2$}{
            perform linear regression to predict the geometric center $(\hat{\bar{i}}_{t})$ using the geometric centers of the previous frames $G$ \;
            shift instance $i \in \hat{\mathcal I}_{x_{t^{\text{last}}}}$ by the vector $\left( \hat{\bar{i}}_{t} - \bar{i}_{t^{\text{last}}} \right)$ and denote the resulting instance by $i^{\text{shift}}$  \;
            $ \hat{\mathcal I}^{\text{detect}}_{x_{t}} := \hat{\mathcal I}^{\text{detect}}_{x_{t}} \cup \{ i^{\text{shift}} \} $\;
        }
    }

  }
  /* check covering  */ \;
  \For{$t = 1,\ldots, T$}{
    \For{$j \in \hat{\mathcal I}^{\text{detect}}_{x_{t}}$}{
        \If{$ O_{i,\mathcal R} < 0.8$ and $\max_{k \in \hat{\mathcal I}_{x_{t}}} \tilde{O}_{j,k} \leq 0.95$}{
            $ \hat{\mathcal I}_{x_{t}} := \hat{\mathcal I}_{x_{t}} \cup \{ j \} $\;
        }
    }
  }
\label{alg:detect}
\end{algorithm}
%
%
\subsection{Metrics}\label{sec:metrics}
In instance segmentation, the neural network provides information of the instances like geometric characteristics or the softmax output. For example, a probability distribution over the classes $y \in \mathcal{C}$ for each pixel $z$ or only for each instance is provided depending on the network architecture. However, a probability distribution is not available for the detected instances, thus we construct metrics based on information that is also obtainable for the detected ones.

First, we define the \emph{size} for an instance $i$ by $S=|i|$ and the size divided into \emph{inner} $S_{in}$ and \emph{boundary} $S_{bd}$. The inner $i_{in}\subset i$ of an instance consists of all pixels whose eight neighboring pixels are also elements of $i$ and therefore, the boundary is $i_{bd}=  i \setminus i_{in}$. The \emph{relative instance sizes} are given by $\tilde S = S/S_{bd}$ and $\tilde S_{in} = S_{in}/S_{bd}$. The separate treatment of inner and boundary is motivated by poor or false predictions that often results in fractal instance shapes, i.e., a relatively large amount of boundary pixels, measurable by the relative measures.

Furthermore, we add the \emph{geometric center} $\bar{i}$ \cref{eq:geom_center} and the predicted \emph{class} $c$ to our set of metrics. In addition to the class, the instance segmentation network provides for each instance a confidence value, also called \emph{score} value denoted by $s$. We calculate the score value for the detected instances as the average score value from the previous frames of the respective instance.

Next, we define the \emph{occlusion} measure $o$ using instances in two consecutive frames. Instance $i$ of frame $t-1$ is shifted such that instance $i$ and its matched counterpart in frame $t$ have a common geometric center. Then, the occlusion of instance $i$ in frame $t$ is given by 
$O_{i,K}$ where $K = \bigcup_{k \in \hat{\mathcal I}_{x_{t}} \setminus i } k$. Large occlusions may indicate poorly predicted instances.

Besides an instance segmentation network, we consider a monocular depth estimation network which use the same input images as the instance segmentation network and can run in parallel. Given an image $x$, the pixel-wise depth prediction is denoted by $D_{z}(x)$, see \Cref{fig:gt_pred} for an example. To obtain metrics per instance, we define the \emph{mean depth} as
\begin{equation}
    \bar D_{*}(i) = \frac{1}{S_{*}} \sum_{z\in i_{*}} D_z(x), \ \ * \in \{ \_,in,bd \}
\end{equation}
where a distinction between inner and boundary is also made. The \emph{relative mean depth} measures are calculated as $\tilde {\bar D} = \bar D \tilde S$ and $\tilde {\bar D}_{in} = \bar D_{in} \tilde S_{in}$. A lower value of mean depth indicates reliable instances, while a higher value suggests uncertainty. 

Moreover, for each instance $i$ in frame $t$ a time series of the mean depth $\bar D$ of the $5$ previous frames is constructed. Using linear regression, the mean depth $\hat{\bar{D}}$ in frame $t$ is predicted if the instance exists in at least two previous frames. The \emph{deviation} between expected mean depth and mean depth $ |\hat{\bar{D}} - \bar D | $ is used as a temporal measure denoted by $d_{d}$. Small deviations $d_{d}$ indicate consistency over time of the instance.

Finally, we add $5$ further measures introduced in \cite{Maag2020} to our set of metrics. Analogous to the depth deviation $d_{d}$, we calculate the deviation of the instance size $d_{s}$ and of the geometric center $d_{c}$. For the survival analysis \cite{Moore2016} metric $v$, time series of metrics are also considered. More precisely, time series of the previously presented single frame metrics serve as input for a Cox regression \cite{Cox1972} to predict the survival time of an instance $i$ in frame $t$. A lower value of $v$ indicates uncertainty. The next metric $r$ is based on the height to width ratio of the instances separated by class. Deviations from this ratio for an instance $i$ suggest false predictions. The final measure $f$ describes the variation of an instance in two consecutive frames by calculating the overlap \cref{eq:overlap}. Poorly predicted instances can result in large deformations.

In summary, we use the following set of metrics 
\begin{align}
    U^{i} = & \ \{ S, S_{in}, S_{bd}, \tilde S, \tilde S_{in} \} \cup \{ \bar D, \bar D_{in}, \bar D_{bd}, \tilde {\bar D}, \tilde {\bar D}_{in} \} \notag \\
    & \cup \{ \bar{i}, c, s, o, d_{d} \} \cup \{  d_{s}, d_{c}, v, r, f \} \, .
\end{align}
%
%
\subsection{Meta classification}\label{sec:meta}
Meta classification is used to identify false positive instances provided by a neural network. A predicted instance is considered as a false positive, if the intersection over union is less than a threshold $h$. The IoU is a commonly used performance measure that quantifies the degree of overlap of prediction and ground truth \cite{Jaccard1912}. If the overlap of a predicted instance $i$ with a ground truth instance $g$ is the highest compared to the other ground truth instances, the IoU is calculated between these two instances $i$ and $g$. Meta classification refers to the task of classifying between IoU $< h$ and IoU $\geq h$ for all predicted instances. In semantic segmentation, classification is typically between IoU $=0$ and IoU $>0$ and in object detection between IoU $< 0.5$ and IoU $\geq 0.5$. 

We perform meta classification using the metrics introduced in \Cref{sec:metrics} as input for the classifier, in particular, we apply time series of these metrics. For an instance $i \in \hat{\mathcal I}_{x_{t}}$ in frame $t$ the metrics $U^{i}_{t}$ are obtained as well as $U^{i}_{t'}$ from previous frames $t' < t$ due to the tracking of the instances. Meta classification is conducted by means of these time series of metrics $U^{i}_{k}$, $k = t-n, \ldots, t$ where $n$ describes the number of considered frames. As classifier model, we use gradient boosting \cite{Friedman2002} that performs best in comparison to linear models and shallow neural networks as shown in \cite{Maag2020}. We study the benefit from using time series and to which extent meta classification along with our false negative detection method can improve the overall network performance compared to the application of a score threshold during inference.
%
%
%
\section{Numerical Results}\label{sec:result}
In this section, we study the properties of the metrics introduced in the previous section and the influence of different lengths of the time series which serve as input to the meta classifier. Furthermore, we evaluate to which extent our detection algorithm fused with uncertainty based meta classification can improve an instance segmentation performance and reduce false negative instances. To this end, we compare this approach with ordinary score thresholds in terms of numbers of false negatives and false positives, i.e., recall and precision rates. 

We perform our tests on two datasets for multi-object tracking and instance segmentation.
The KITTI dataset \cite{Geiger2012} contains $21$ street scene image sequences from Karlsruhe (Germany) consisting of $8,\!008$ $1,\!242 \times 375$ images. The MOT dataset \cite{Milan2016} containing $2,\!862$ images with resolutions of $1,\!920 \times 1,\!080$ ($3$ image sequences) and $640 \times 480$ ($1$ image sequence) provides scenes from pedestrian areas and shopping malls. For both datasets annotated image sequences, i.e., tracking IDs and instance segmentations, are available \cite{Voigtlaender2019}. The KITTI dataset includes labels for the classes car and pedestrian, while the MOT dataset only contains labels for the class pedestrian.

In our experiments, we consider the Mask R-CNN \cite{He2017} and the YOLACT network \cite{Bolya2019}.
The YOLACT network is designed for a single GPU, therefore the underlying architecture is slim. For training this network, we use a ResNet-50 \cite{He2015} as backbone and pre-trained weights for ImageNet \cite{Russakovsky2015}. We choose $12$ image sequences from the KITTI dataset consisting of $5,\!027$ images and $300$ images (extracted from $2$ sequences) from the MOT dataset for training. The remaining $9$ sequences of the KITTI dataset serve as validation set (same splitting as in \cite{Voigtlaender2019}). We achieve a mean average precision (mAP) \cite{Everingham2009} of $57.15\%$. Since we are using $300$ images of the MOT dataset for training, we split $2$ of the $4$ sequences into train/validation and remove $90$ frames (equal to $3$ seconds) of each at this splitting point due to redundancy in image sequences. We achieve a mAP of $52.37\%$ on the remaining $2,\!382$ images of the MOT dataset.
The Mask R-CNN is focused on high-quality instance-wise segmentation masks representations. For training, we use a ResNet-101 as backbone and start from weights for COCO \cite{Lin2014}. As training set, we choose the same $12$ image sequences of the KITTI dataset and the validation sets also remain the same. For the KITTI dataset, we obtain a mAP of $89.79\%$ and for the MOT dataset of $78.99\%$.
During training of both networks, the validation set is neither used for early stopping nor for parameter tuning.

For our depth metrics extracted from a depth estimation, we use the network introduced in \cite{Lee2019}. This network utilizes local planar guidance layers at multiple stages of the decoding phase for an effective guidance of densely encoded features. As backbone, we consider the DenseNet-161 \cite{Huang2017} and pre-trained weights for ImageNet. We train this network over a maximum of $50$ epochs on $20,\!750$ RGB images of the KITTI dataset where annotated depth maps are available. The model which achieves the lowest scale invariant logarithmic error (silog) \cite{Eigen2014} on $672$ validation images is used. This validation silog amounts to $8.493$. We also evaluate on the $9$ images sequences of the KITTI dataset, that we consider for further experiments, achieving a test silog of $9.459$ on these $2,\!891$ images (for $90$ images is no depth ground truth available).

During inference, a score threshold is used to remove instances with low score values followed by a non-maximum suppression to avoid multiple predictions for the same instance. Since we compare our detection method with the application of different score values, we apply none or a very low score threshold during the inference. For the KITTI dataset, we choose a score threshold of $0$ in both networks to use all predicted instances for further experiments. For the MOT dataset, we choose relatively low score thresholds, i.e., $0.3$ for the Mask R-CNN and $0.05$ for the YOLACT network. In our experiments, we use $9$ images sequences of the KITTI dataset consisting of $2,\!981$ images and $4$ sequences of the MOT dataset containing $2,\!382$ images (equal to the validation sets used for instance segmentation).
%
%
\subsection{Meta Classification}
First, we study the predictive power of the instance-wise metrics which serve as input for the meta classifier by computing the Pearson correlation coefficients $\rho$ between selected metrics of $U^{i}$ and the IoU, see \Cref{tab:corr}.
\begin{table}[t]
\centering
\caption{A selection of correlation coefficients $\rho$ with respect to IoU.}
\label{tab:corr}
\scalebox{0.915}{
\begin{tabular}{c|cc|cc}
\cline{1-5}
\multicolumn{1}{c}{} & \multicolumn{2}{c|}{KITTI} & \multicolumn{2}{c}{MOT} \rule{0mm}{2.5mm}\\
\multicolumn{1}{c}{} & Mask R-CNN & YOLACT & Mask R-CNN & YOLACT \\
\cline{1-5}
$\tilde S_{in}$ & $0.4232$ & $0.2869$ & $0.5144$ & $0.6975$ \rule{0mm}{3.5mm}\\
$\tilde {\bar D}_{in}$ & $0.5664$ & $0.3812$ & $0.5906$ & $0.6897$ \rule{0mm}{3.0mm}\\
$s$ & $0.6490$ & $0.7936$ & $0.6611$ & $0.6770$ \rule{0mm}{3.0mm}\\
$o$ & $-0.0417$ & $0.2960$ & $-0.1598$ & $-0.2177$ \rule{0mm}{2.5mm}\\
$d_{d}$ & $-0.1960$ & $-0.1611$ & $-0.0405$ & $-0.1363$ \rule{0mm}{2.5mm}\\
\cline{1-5}
\end{tabular} }
\end{table}
For both networks and both datasets, the score value $s$ shows a strong correlation with the IoU. The relative instance size $\tilde S_{in}$ as well as the relative mean depth $\tilde {\bar D}_{in}$ demonstrate high correlations for the MOT dataset (for both datasets) and the KITTI dataset in combination with the Mask R-CNN.
For meta classification (false positive detection: IoU $<h$ vs.\ IoU $\geq h$), we use the set of metrics $U^{i}$ as input and investigate the influence of time series. We present this set of metrics $U^{i}_{t}$ of a single frame $t$ to the meta classifier and then, the metrics are extended to time series with a length of up to $10$ previous time steps $U^{i}_{k}$, $k=t-10, \ldots, t-1$. Furthermore, we use a (meta) train/val/test splitting of $70\%/10\%/20\%$ and average the results over $10$ runs for the KITTI dataset and $4$ runs for the MOT dataset by randomly sampling this splitting. For the presented results, we choose an IoU threshold of $h = 0.5$.

We apply our detection algorithm to the predictions of both instance segmentation networks for both datasets. The corresponding numbers of predicted instances and instances obtained by the detection algorithm are provided in \Cref{tab:num}.
\begin{table}[t]
\centering
\caption{Number of instances predicted by a neural network (PI) and number of detected instances (DI) using \Cref{alg:detect}.}
\label{tab:num}
\scalebox{0.915}{
\begin{tabular}{c|cc|cc}
\cline{1-5}
\multicolumn{1}{c}{} & \multicolumn{2}{c|}{KITTI} & \multicolumn{2}{c}{MOT} \rule{0mm}{2.5mm}\\
\multicolumn{1}{c}{} & Mask R-CNN & YOLACT & Mask R-CNN & YOLACT \\
\cline{1-5}
PI & $19,\!239$ & $25,\!743$ & $36,\!003$ & $22,\!495$ \\
DI & $11,\!787$ & $14,\!694$ & $27,\!374$ & $16,\!400$ \\
\cline{1-5}
\end{tabular} }
\end{table}
For the KITTI dataset, we obtain $31,\!026$ instances (not yet matched over time) for the Mask R-CNN of which $20,\!337$ have an IoU $< 0.5$ and for the YOLACT network out of $40,\!437$ instances, $32,\!114$ have an IoU $< 0.5$. For the MOT dataset, this ratio is $63,\!377 / 45,\!307$ for the Mask R-CNN and $38,\!895 / 26,\!153$ for the YOLACT network. The relatively high number of false positive instances is the motivation to perform meta classification after the detection algorithm in order to get rid of false positives. As performance measures for the meta classification, we consider the classification accuracy and $\auroc$. The best results are given in \Cref{tab:meta_classif}.
\begin{table}[t]
\centering
\caption{Meta classification results (with corresponding standard deviations) using an IoU threshold $h=0.5$. The super script denotes the number of considered frames where the best performance and in particular the given values are reached.}
\label{tab:meta_classif}
\scalebox{0.915}{
\begin{tabular}{ccc}
\cline{1-3}
\multicolumn{1}{c}{} & $\acc$ & $\auroc$ \\
\cline{1-3}
KITTI \& Mask R-CNN & $95.64\%\pm0.74\%^7$ & $99.04\%\pm0.31\%^5$ \rule{0mm}{3.5mm}\\
KITTI \& YOLACT & $96.22\%\pm1.15\%^7$ & $99.30\%\pm0.34\%^7$ \\
\cline{1-3}
MOT \& Mask R-CNN & $93.60\%\pm3.42\%^9$ & $98.25\%\pm0.93\%^9$ \rule{0mm}{3.5mm}\\
MOT \& YOLACT & $87.66\%\pm5.45\%^9$ & $96.22\%\pm1.22\%^5$ \\
\cline{1-3}
\end{tabular} }
\end{table}
We achieve $\auroc$ values between $96.22\%$ and $99.30\%$. Furthermore, we notice that gradient boosting benefits from temporal information which can also be observed in \Cref{fig:results} (left) where the $\auroc$ results as functions of the number of frames are given for the KITTI dataset. 
\begin{figure*}[t]
\center
    \subfigure{\includegraphics[scale=0.30]{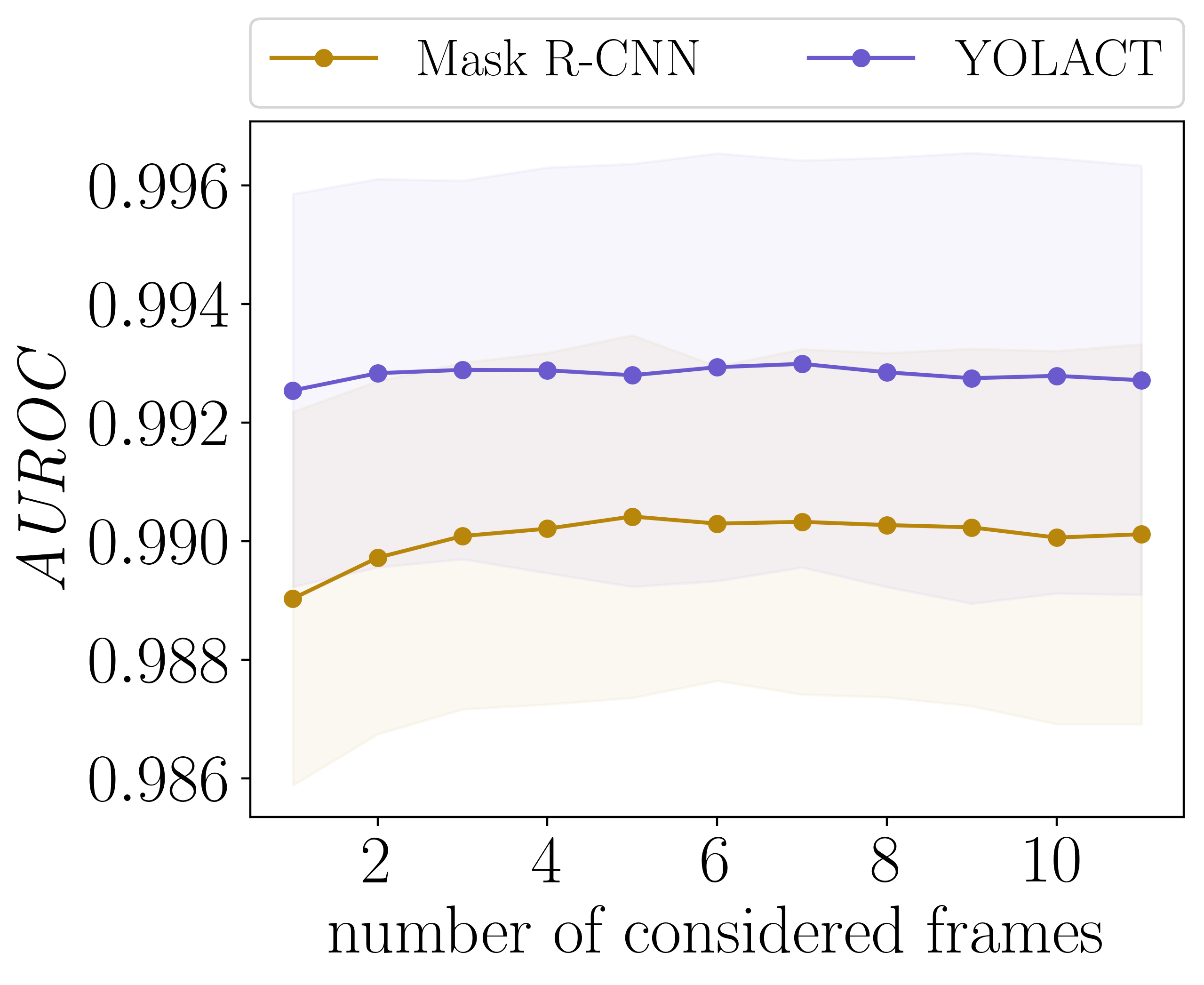}}
    \subfigure{\includegraphics[scale=0.34]{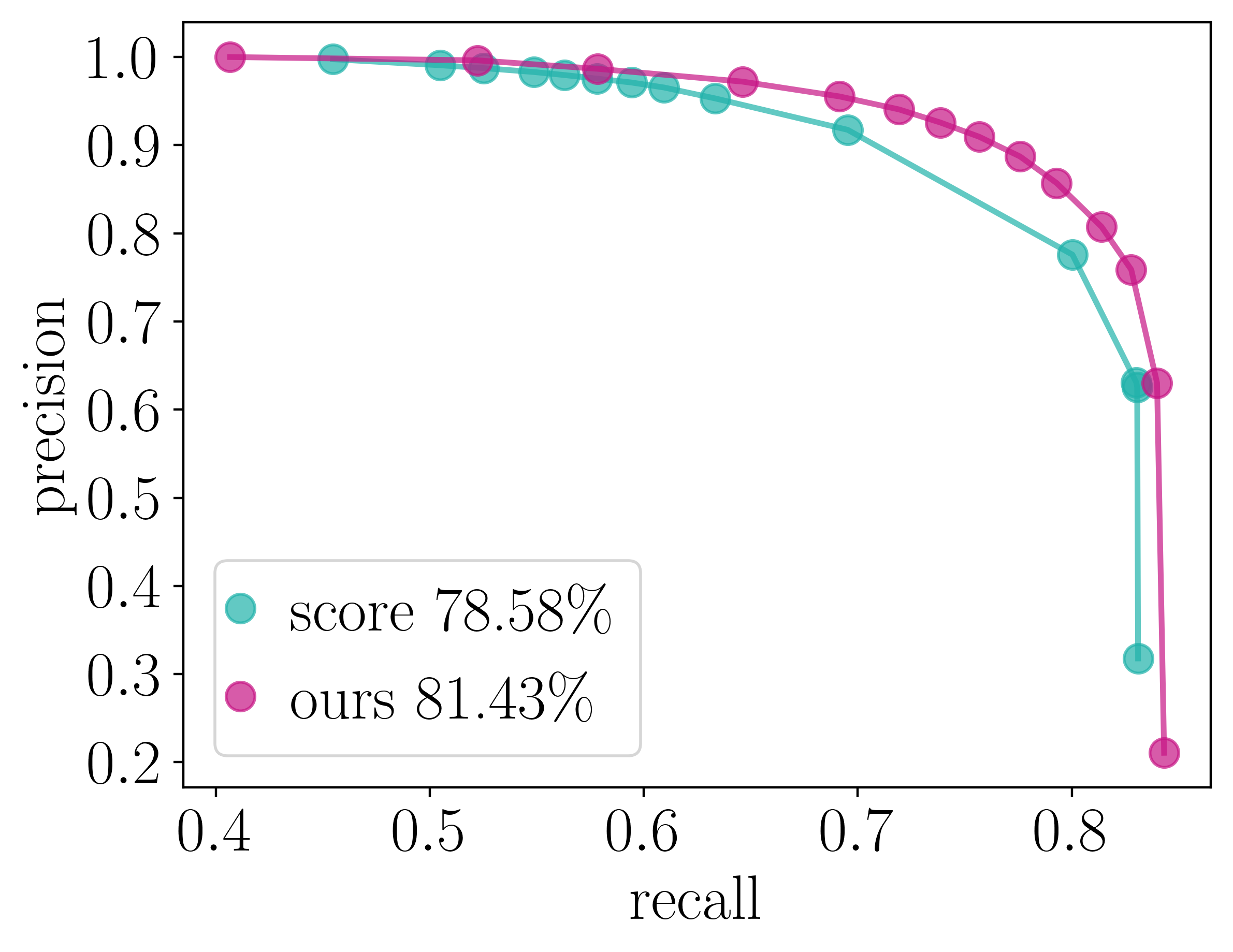}}
    \subfigure{\includegraphics[scale=0.34]{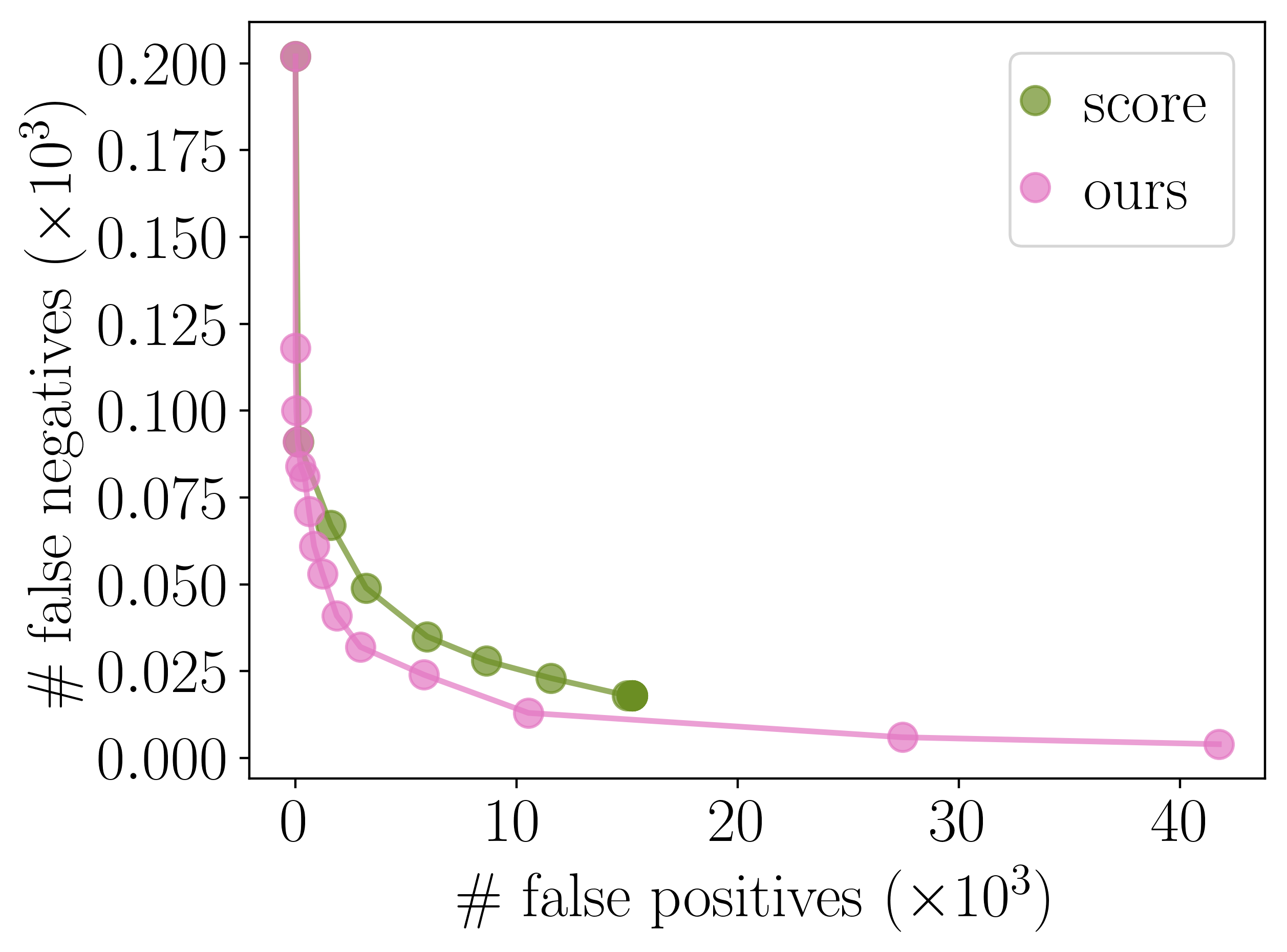}}
    \caption{\emph{Left}: Results for meta classification $\auroc$ as functions of the number of frames for the KITTI dataset. \emph{Center}: Precision-recall curves for an IoU threshold $h = 0.5$ and different score as well as meta classification thresholds for the YOLACT network and the KITTI dataset. \emph{Right}: Number of false positive vs.\ false negative instances for different score and meta classification thresholds for the MOT dataset and the Mask R-CNN network. An IoU threshold of $0$ is used as well as an occlusion level of $0.5<$ IoU$_{bb}$ $\leq 0.6$.}
    \label{fig:results}
\end{figure*}
For further experiments, we use the number of considered frames where the best $\auroc$ performance is reached, respectively.
%
%
\subsection{Evaluation of our Detection Method}
Our detection method assume the instance segmentation by a neural network and tracking IDs. We consider the tracking algorithm for instances introduced in \cite{Maag2020} where the tracking performance is measured by different object tracking metrics. In the following, we compute a few object tracking metrics and apply these to the instances predicted by a network and the additional instances obtained by our detection method. We denote by $\mathit{GT}$ all ground truth instances of an image sequence which are identified by different tracking IDs and following \cite{Milan2016}, we divide these into three cases. An instance is called mostly tracked ($\mathit{MT}$) if it is tracked for at least $80\%$ of frames (out of the total number of frames in which it occurs), mostly lost ($\mathit{ML}$) if it is tracked for less than $20\%$, otherwise partially tracked ($\mathit{PT}$). In addition, we define the number of switches between matched and non-matched $\mathit{smn}_{t}$, i.e., a switch appears when a ground truth instance is matched with a predicted instance in frame $t$ and non-matched in frame $t+1$ and vice versa. The results are shown in \Cref{tab:tracking}.
\begin{table}[t]
\centering
\caption{Object tracking results for the instances obtained by our detection method (including those predicted by an instance segmentation network). The values in brackets correspond to the outcomes for the predicted instances without using the detection step.}
\label{tab:tracking}
\scalebox{0.915}{
\begin{tabular}{c|cc|cc}
\cline{1-5}
\multicolumn{1}{c}{} & \multicolumn{2}{c|}{KITTI} & \multicolumn{2}{c}{MOT} \rule{0mm}{2.5mm}\\
\multicolumn{1}{c}{} & Mask R-CNN & YOLACT & Mask R-CNN & YOLACT \\
\cline{1-5}
$\mathit{GT}$ & $219$ & $219$ & $201$ & $201$ \\
$\mathit{MT}$ & $205$ ($204$) & $129$ ($124$) & $123$ ($120$) & $45$ ($42$)\\
$\mathit{PT}$ & $12$ ($13$) & $71$ ($75$) & $72$ ($75$) & $92$ ($92$)\\
$\mathit{ML}$ & $2$ ($2$) & $19$ ($20$) & $6$ ($6$) & $64$ ($67$)\\
$\mathit{smn}$ & $186$ ($189$) & $352$ ($379$) & $797$ ($824$) & $552$ ($597$)\\
\cline{1-5}
\end{tabular} }
\end{table}
We observe that our algorithm can increase the number of mostly tracked ground truth instances and reduce the number of switches between matched and non-matched for both networks and datasets. Note, the Mask R-CNN mainly achieves better results than YOLACT which can be explained by Mask R-CNN achieving higher mAP values than YOLACT.

The score value describes the confidence of the network's prediction and is chosen during inference to balance the number of false negatives and false positives. We select $15$ different score thresholds between $0$ and $1$ to study the network's detection performance while varying this threshold. We apply our false negative detection algorithm to the predicted instances and fuse this with uncertainty based meta classification. Meta classification provides a probability of observing a false positive instance and thus, we also threshold on this probability with $15$ different values. In our tests, we consider $6$ different IoU thresholds $h$ for meta classification. We feed the meta classifier with all metrics $U^{i}$ including the number of previous frames given in \Cref{tab:meta_classif} independent of the IoU threshold. We compare ordinary score thresholding with our method using the area under precision-recall curve (AUC) as performance measure. As our detection algorithm receives time series of predicted instances as input, certain ground truth instances cannot be found. For this reason, we exclude ground truth instances for further testing if the respective instance defined by its tracking ID is never found by the instance segmentation network and in the frames before the instance is first detected by the network. This number depends on the respective IoU threshold, i.e., at higher values, more ground truth instances are not detected. For the Mask R-CNN, at most $2.19\%$ and for the YOLACT network, $17.62\%$ of the ground truth instances are not found and for this reason, not evaluated. The $\auc$ results are given in \Cref{tab:auc}.
\begin{table}[t]
\centering
\caption{$\auc$ results for score thresholding vs.\ our false negative detection algorithm fused with meta classification thresholding.}
\label{tab:auc}
\scalebox{0.915}{
\begin{tabular}{c|cc|cc}
\cline{1-5}
\multicolumn{1}{c}{KITTI $\&$} & \multicolumn{2}{c}{Mask R-CNN} & \multicolumn{2}{c}{YOLACT} \rule{0mm}{2.5mm}\\
\cline{1-5}
IoU threshold $h$ & score & ours & score & ours \\
\cline{1-5}
$0.5$ & $91.17\%$ & $\mathbf{92.35}\boldsymbol{\%}$ & $78.58\%$ & $\mathbf{81.43}\boldsymbol{\%}$ \\
$0.4$ & $93.19\%$ & $\mathbf{93.81}\boldsymbol{\%}$ & $81.03\%$ & $\mathbf{84.17}\boldsymbol{\%}$ \\
$0.3$ & $93.98\%$ & $\mathbf{94.45}\boldsymbol{\%}$ & $82.28\%$ & $\mathbf{85.83}\boldsymbol{\%}$ \\
$0.2$ & $94.49\%$ & $\mathbf{94.88}\boldsymbol{\%}$ & $81.83\%$ & $\mathbf{85.61}\boldsymbol{\%}$ \\
$0.1$ & $94.67\%$ & $\mathbf{95.08}\boldsymbol{\%}$ & $82.19\%$ & $\mathbf{86.06}\boldsymbol{\%}$ \\
$0.0$ & $94.99\%$ & $\mathbf{95.39}\boldsymbol{\%}$ & $83.87\%$ & $\mathbf{87.78}\boldsymbol{\%}$ \\
\cline{1-5}
\multicolumn{5}{c}{} \\
\cline{1-5}
\multicolumn{1}{c}{MOT $\&$} & \multicolumn{2}{c}{Mask R-CNN} & \multicolumn{2}{c}{YOLACT} \rule{0mm}{2.5mm}\\
\cline{1-5}
IoU threshold $h$ & score & ours & score & ours \\
\cline{1-5}
$0.5$ & $77.45\%$ & $\mathbf{79.05}\boldsymbol{\%}$ & $62.32\%$ & $\mathbf{63.73}\boldsymbol{\%}$ \\
$0.4$ & $82.44\%$ & $\mathbf{82.76}\boldsymbol{\%}$ & $68.39\%$ & $\mathbf{69.71}\boldsymbol{\%}$ \\
$0.3$ & $\mathbf{85.17}\boldsymbol{\%}$ & $85.02\%$ & $71.29\%$ & $\mathbf{72.77}\boldsymbol{\%}$ \\
$0.2$ & $\mathbf{86.94}\boldsymbol{\%}$ & $86.85\%$ & $72.31\%$ & $\mathbf{74.22}\boldsymbol{\%}$ \\
$0.1$ & $\mathbf{88.56}\boldsymbol{\%}$ & $88.43\%$ & $73.89\%$ & $\mathbf{76.96}\boldsymbol{\%}$ \\
$0.0$ & $\mathbf{89.85}\boldsymbol{\%}$ & $89.48\%$ & $76.11\%$ & $\mathbf{81.77}\boldsymbol{\%}$ \\
\cline{1-5}
\end{tabular} }
\end{table}
Smaller IoU thresholds increase the possibility of matches between ground truth and predicted instances and consequently the $\auc$ value increases. We observe that our method performs better in most cases compared to the use of a score threshold (or obtains very similar values). We obtain $\auc$ values up to $95.39\%$. An example for the precision-recall curves is shown in \Cref{fig:results} (center) using an IoU threshold of $h=0.5$ for the KITTI dataset and the YOLACT network. Each point represents one of the chosen score or meta classification thresholds. Our detection method achieves a lower number of errors, i.e., higher recall and precision rates. In particular, we can reduce the number of false negative instances.

For further experiments, we divide the ground truth instances in different occlusion levels. To this end, we calculate for each instance the IoU$_{bb}$ with the other ones represented as bounding boxes. On the one hand, for high IoU$_{bb}$ values, the instance can be partially covered by other instances or even cover others. On the other hand, for low IoU$_{bb}$ values, detecting the ground truth instance through an instance segmentation network is more simple as if the instance is occluded and located in a crowd of instances. In \Cref{fig:results} (left), the comparison of ordinary score thresholding and our detection method is stated in terms of the number of remaining false negatives and false positives considering an occlusion level of $0.5 <$ IoU$_{bb} \leq 0.6$. As before, each point represents one of the chosen thresholds. We achieve a lower number of false negative and false positive instances. This error reduction is also reflected in the $\auc$ values. The $\auc$ value obtained by score thresholding is $98.93\%$ while the detection algorithm fused with meta classification achieves $99.86\%$. This example shows that our method can improve the instance segmentation also in more difficult cases, i.e., at higher occlusion levels.

In \Cref{fig:timeline}, an example of how our detection algorithm works is demonstrated for a ground truth instance of class car and the KITTI dataset.
\begin{figure}[t]
\center
    \subfigure{\includegraphics[trim=3 0 0 0,clip,width=0.27\textwidth]{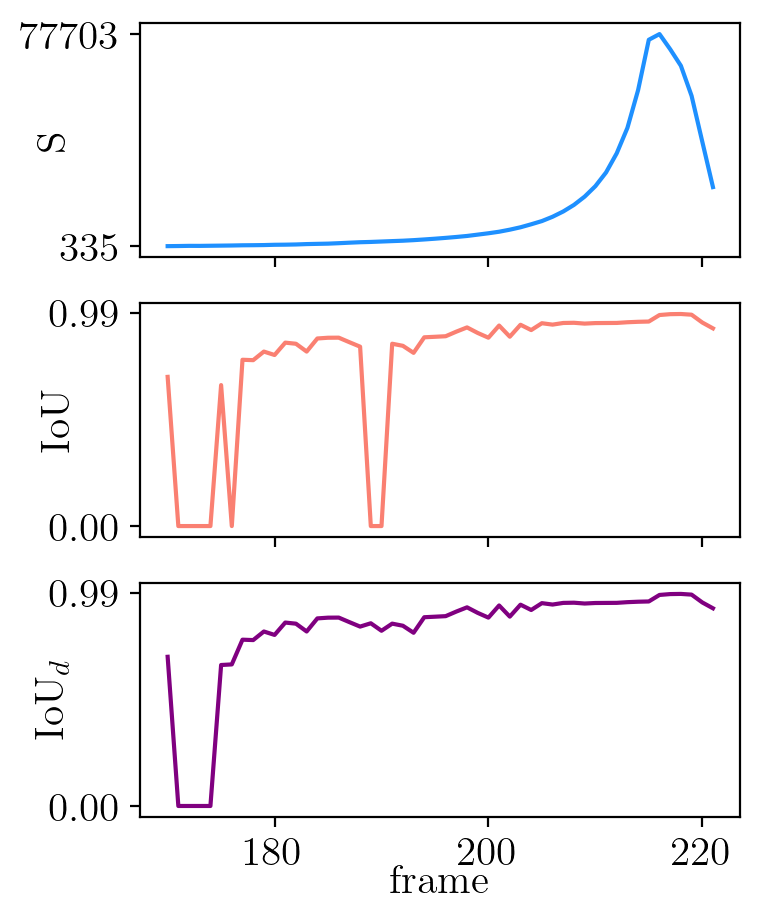}}
    \subfigure{\includegraphics[trim=40 313 40 0,clip,scale=0.34]{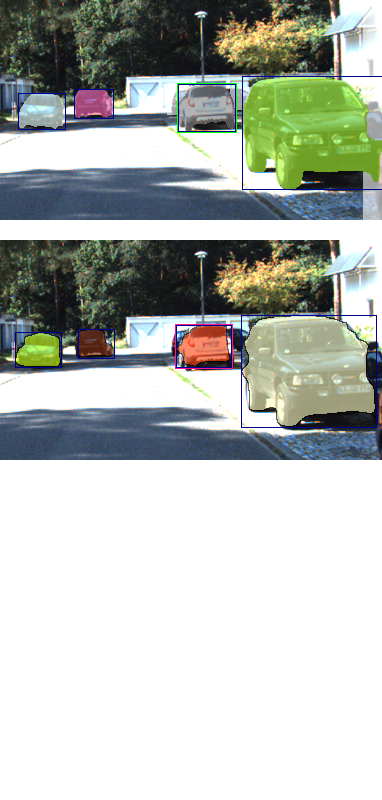}}
    \caption{\emph{Left}: Time series of size $S$ for a ground truth instance of the KITTI dataset, the calculated IoU between this ground truth instance and instances predicted by the YOLACT network as well as the IoU$_{d}$ after the application of our detection method. \emph{Top right}: Corresponding ground truth image in frame $190$ including the considered instance (bounded by a green box). \emph{Bottom right}: Instance segmentation followed by our detection algorithm which constructs the car instance with the pink bounding box.}
    \label{fig:timeline}
\end{figure}
The time series of the instance size $S$ and of the IoU between this ground truth instance and instances obtained by the YOLACT network are shown. We observe $3$ areas where the instance has an IoU $= 0$ and hence, has not been detected by the network. Our detection algorithm is applied to the predicted instances which results in the time series of IoU$_{d}$. Since the procedure requires at least two geometric centers of an instance to generate further instances, the first plateau cannot be fixed, although the following two ones can be. Both images, ground truth (top) and instance segmentation (bottom), represent frame $190$ where the instance (bounded by a green box) is not detected by the network, but by our method. We construct the car labeled with the pink bounding box in the segmentation image. Noteworthy, though no score threshold is used during inference, the network has not predicted any instance at this position. Also note, both images correspond to the zoomed image sections and the ground truth instance is relatively small which our algorithm can handle. Our false negative reduction method consisting of the detection algorithm and meta classification can be applied to any instance segmentation network after training and thus, does not increase the network complexity.
%
%
\section{Conclusion}\label{sec:concl}
In this work, we proposed a post-processing method applicable to any instance segmentation network to reduce the number of false negative instances by performing a detection and a false positive pruning step. Our light-weight detection algorithm is based on inconsistencies in the time series of tracked instances and constructs new instances that the neural network may have missed using the information of previous frames. Since the number of instances can be greatly increased, we deployed meta classification to reduce false positive instances. As input for the meta classification model, instance-wise metrics were constructed characterizing uncertainty, geometry and depth of a given instance. We studied the properties of the metrics and the influence of different time series lengths on the meta classification model. We achieved $\auroc$ values of up to $99.30\%$. In our tests, we compared our approach with the application of a score threshold during inference and improved the networks' prediction accuracy. We obtained area under precision-recall curve values of up to $95.39\%$. In particular, the number of false negative instances can be reduced.

\section*{Acknowledgment}
We thank Hanno Gottschalk and Matthias Rottmann for discussion and useful advice.

\bibliographystyle{IEEEtran}
\bibliography{biblio}

\end{document}